\documentclass{article}

\usepackage{microtype}
\usepackage{graphicx}
\usepackage{caption}
\usepackage{subcaption}
\usepackage{booktabs} % for professional tables

\usepackage{hyperref}

\usepackage[accepted]{icml2025}

\usepackage{amsmath}
\usepackage{amssymb}
\usepackage{mathtools}
\usepackage{amsthm}
\usepackage{relsize}

\usepackage[capitalize,noabbrev]{cleveref}

\theoremstyle{plain}

\theoremstyle{definition}

\theoremstyle{remark}

\usepackage[textsize=tiny]{todonotes}

\icmltitlerunning{Guided Generation for Developable Antibodies}

\begin{document}

\twocolumn[
\icmltitle{Guided Generation for Developable Antibodies}




\begin{icmlauthorlist}  
\icmlauthor{Siqi Zhao}{comp}
\icmlauthor{Joshua Moller}{comp}
\icmlauthor{Porfirio Quintero-Cadena}{comp}
\icmlauthor{Lood van Niekerk}{comp}
\end{icmlauthorlist}

\icmlaffiliation{comp}{Ginkgo Bioworks, Boston, USA}
% Corresponding: Ammar,Rich,Siqi/Josh
\icmlcorrespondingauthor{Joshua Moller}{jmoller@ginkgobioworks.com}
\icmlcorrespondingauthor{Lood van Niekerk}{lvanniekerk@ginkgobioworks.com}

\icmlkeywords{Machine Learning, ICML, antibody, generative, discrete diffusion, guidance, SVDD}

\vskip 0.3in
]



\printAffiliationsAndNotice{}  % leave blank if no need to mention equal contribution

%%%%%%%%%%%%%%%%%%%%%%%%%%%
% NOTES of things to do for camera-ready
% Get TAP predictions for all generated sequences
% Add descriptions of each property
% Appendix additions:
% Porfi comment on evaluation and SSN details, and later on the sequence identity specifics
% Experimental extensions: 
% Not sure, could benchmark with NOS?
%%%%%%%%%%%%%%%%%%%%%%%%%%%

\begin{abstract}
Therapeutic antibodies require not only high-affinity target engagement, but also favorable manufacturability, stability, and safety profiles for clinical effectiveness. These properties are collectively called `developability'. To enable a computational framework for optimizing antibody sequences for favorable developability, we introduce a guided discrete diffusion model trained on natural paired heavy- and light-chain sequences from the Observed Antibody Space (OAS)~\cite{Olsen2022-so} and quantitative developability measurements for 246 clinical-stage antibodies. To steer generation toward biophysically viable candidates, we integrate a Soft Value-based Decoding in Diffusion (SVDD) Module that biases sampling without compromising naturalness. In unconstrained sampling, our model reproduces global features of both the natural repertoire and approved therapeutics, and under SVDD guidance we achieve significant enrichment in predicted developability scores over unguided baselines. When combined with high-throughput developability assays, this framework enables an iterative, ML-driven pipeline for designing antibodies that satisfy binding and biophysical criteria in tandem.
\end{abstract}

\section{Introduction}

Therapeutic antibodies are pivotal biomolecules with applications spanning oncology \cite{Paul2024-gc}, autoimmune disorders \cite{Chan2010-ez}, infectious diseases \cite{Sparrow2017-rm}, and metabolic conditions \cite{Lu2020-ui}. Beyond high-affinity target binding, a developable antibody must also exhibit favorable manufacturability, formulation stability, and safety profiles to support scalable production and reliable delivery \cite{Jain2017-wj,Carter2022-gn}. Although high-throughput screens and \textit{in silico} tools can identify and design candidates for binding affinity \cite{Agarwal2024-cu,Frey2025-ir}, comprehensive machine-driven frameworks for optimizing key developability attributes are still lacking. There is therefore an urgent need for computational methods that not only predict developability properties to triage high-quality binders for downstream validation, but also guide the redesign of existing antibody sequences toward improved developability.

Contemporary \textit{in silico} approaches typically benchmark candidate antibodies' biophysical attributes against those of clinically approved therapeutics \cite{Raybould2019-bs, Raybould2024-df, Park2024-dy}. However, these comparative metrics often overlook the inherent variability within approved antibody repertoires and, in the absence of true negative controls, cannot establish meaningful developability thresholds. Moreover, because experimental workflows almost invariably screen solely for target binding -- omitting parallel assessments of manufacturability, solubility, or stability -- there is a critical need for computational methods that optimize sequences for favorable developability properties. Other model-based optimization frameworks have been proposed \cite{Sinai2020-iq, Stanton2022-bk, Gruver2023-wk, Reddy2024-lm}, and some specific to antibodies. Recently, another study \cite{Wang2025-sq} demonstrated the usefulness of using physical descriptors derived from predicted structures from protein language models to guide antibody design.

As a step towards addressing this challenge, we developed a machine learning-guided generative framework anchored on our newly published developability dataset. We trained our generative model on natural paired heavy- and light-chain sequences from the Observed Antibody Space (OAS) and built quantitative regressors using the comprehensive developability measurements reported in \cite{Arsiwala2025-fz} for 246 antibodies spanning clinical use and trial stages. We take a complementary approach to directly use experimentally generated developability measurements from diverse clinically approved antibodies to guide antibody design. We show that, without any conditional input, our model can produce novel sequences that mirror both natural repertoire diversity and features of clinically approved antibodies. Moreover, by integrating a Soft Value-based Decoding in Diffusion (SVDD) \cite{Li2024-lr} guidance module, we can bias generation toward candidates with predicted favorable developability. By employing a derivative-free guidance approach, we established a flexible framework that is compatible with various types of predictors. When paired with high-throughput automated assays, this framework offers a powerful avenue to design therapeutic antibodies that meet both affinity and biophysical criteria.

% Related works section or part of the introduction:
% Guidance is cool, can do DPO, IRPO, classifier-free, SVDD, whatever the one was that we used earlier, and some other people have also done optimization for properties in the context of antibody discrete diffusion.
% LaMBO Stanton2022: https://proceedings.mlr.press/v162/stanton22a/stanton22a.pdf
% NOS Gruver2023 https://openreview.net/forum?id=MfiK69Ga6p

\begin{figure}
    \centering
    \includegraphics[width=0.8\linewidth]{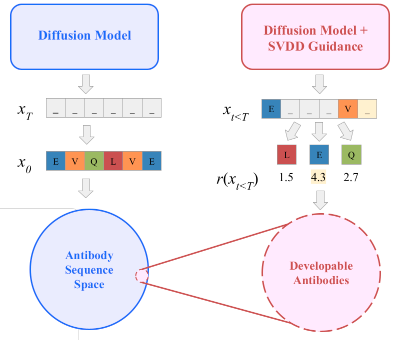}
    \caption{\textbf{Overview of our generative framework for novel antibodies.} This framework incorporates an ESM2-based diffusion model trained with paired antibodies sequences from OAS. For guided generation, soft value-based decoding in diffusion (SVDD) was used with developability predictors trained with data from \cite{Arsiwala2025-fz}.  }
    \label{fig:svdd_diagram}
\end{figure}
\subsection{Generative model}  % Lood: can move some of these details to the supplement, like the clustering

We trained an antibody-specific masked discrete diffusion model on the Observed Antibody Space (OAS) database of sequenced antibodies from over 80 studies. Specifically, we used the order-agnostic diffusion model (OADM)~\cite{Hoogeboom2021-ig} training objective (which had previously been used in EvoDiff~\cite{Alamdari2023-mt}). For simplicity, we re-trained an ESM-2 (8M)~\cite{Lin2023-ss} architecture using this objective.
The OAS dataset contains $\sim$2.4 billion unpaired and $\sim$1.8 million paired sequences collected from human B-cell sequencing. 
We used the AntiRef-90~\cite{Briney2023-dc} version of OAS to remove partial sequences and clustered using MMseqs2~\cite{Steinegger2017-rr} at a 90\% threshold which yielded 1.68 million clusters. % Can show our figures/OAS_paired_clusters here

During training we concatenated the paired heavy and light chains into a single sequence \texttt{<heavy>|<light>} by introducing a pipe token (\texttt{|}). During generation, we sampled a sequence length for the concatenated heavy and light chains from the training dataset and generated sequences according to that length.
Usually order-agnostic models are decoded in random orders, but we found that decoding according to minimum entropy positions yielded better sequences. We also experimented with different sampling temperatures in the softmax function according to the formula $p(x_i) = \frac{e^{\frac{x_i}{T}}}{\mathlarger{\sum}_{j=1}^D{e^{\frac{x_j}{T}}}}$. Identically to language models, higher temperatures correspond to more diverse generations and lower temperatures approach greedy/deterministic samples for each position.

\subsection{Developability dataset}
Recently, we released a dataset that contains biophysical assay measurements for 9 antibody developability properties across 246 clinical antibodies~\cite{Arsiwala2025-fz}\footnote{Dataset available at \url{https://huggingface.co/datasets/ginkgo-datapoints/GDPa1}}, greatly expanding on the set of 137 clinical antibodies provided in \cite{Jain2017-wj}, and running these assays at higher throughput. The 2017 dataset catalyzed the development of many developability predictors, so we sought to train new predictors on the bigger dataset and demonstrate whether we could apply SVDD guidance using these predictors as oracles.

\subsection{Predictive model}
As a simple oracle, we trained ridge regression models on top of ESM-2 embeddings to predict antibody developability properties using the aforementioned dataset. Heavy and light chain sequences were passed independently to ESM to generate mean-pooled embeddings, which were then concatenated and standardized. Finally, a ridge regression with alpha=0.1 was used.
To evaluate performance, we performed hierarchical clustering to separate the 246 sequences into 5 roughly-equal sized folds maximally separated by pairwise sequence identity, and provide the Spearman and Pearson correlation statistics in Table~\ref{tab:predictor_performance}.

For this work, we focus on two biophysical properties when training our predictors: hydrophobicity (measured by hydrophobic interaction chromatography, HIC) and self-association (measured by affinity capture self-interacting nanoparticle spectroscopy, AC-SINS, at pH 7.4). We selected these properties for two reasons. First, they directly impact the administrability of candidate antibodies. Second, they present a realistic scenario in which predictors suffer from limited performance due to data scarcity, allowing us to assess whether guidance compromises the naturalness of sequence generation.

% Can also put it in a separate file
% \input{tables/table_predictor_performance.tex}
% TODO Sebastien suggestion for camera-ready: Add a column indicating the property measured by each assay. Might need to cut down text width or just list these properties (listed in the preprint under "Developability Properties Cluster in Alignment with Previous Findings")
\begin{table}[h]
    \centering
    \begin{tabular}{lll}
     \toprule
     \textbf{Assay} & \textbf{Spearman's $\rho$} & \textbf{Pearson's R}\\
     \midrule
     % Viscosity (AC-SINS pH7.4) & 0.500 $\pm$ 0.09 & 0.479 $\pm$ 1.12\\
     % Hydrophobicity (HIC RT) & 0.405 $\pm$ 0.09 & 0.360 $\pm$ 0.08\\
    HAC RT & $0.74 \pm 0.22$ & $0.80 \pm 0.24$ \\
    SEC \%Monomer & $0.54 \pm 0.05$ & $0.81 \pm 0.14$ \\
    \textbf{AC-SINS pH 7.4} & $0.49 \pm 0.09$ & $0.49 \pm 0.12$ \\
    \textbf{HIC RT} & $0.42 \pm 0.09$ & $0.34 \pm 0.08$ \\
    CHO PR Score & $0.41 \pm 0.09$ & $0.41 \pm 0.12$ \\
    Ova PR Score & $0.40 \pm 0.09$ & $0.28 \pm 0.17$ \\
    AC-SINS pH 6.0 & $0.38 \pm 0.09$ & $0.33 \pm 0.17$ \\
    Tm1 & $0.35 \pm 0.16$ & $0.32 \pm 0.19$ \\
    Tm2 & $0.20 \pm 0.14$ & $0.33 \pm 0.17$ \\
    SMAC RT & $0.18 \pm 0.17$ & $0.00 \pm 0.14$ \\
    Titer & $0.17 \pm 0.14$ & $0.17 \pm 0.19$ \\
    Purity \%LC+HC & $0.10 \pm 0.21$ & $0.11 \pm 0.18$ \\
    \bottomrule
    \end{tabular}
    \caption{Results of training an ESM-2 embedding-based regression models (oracles) on all of the properties in the dataset, as assessed by the average Spearman and Pearson correlations across 5 folds of cross-validation. Assays are ordered by Spearman rank correlation coefficient. The two chosen properties used for guidance in this work are denoted by bold.}
    \label{tab:predictor_performance}
\end{table}

% Figure on generation
\begin{figure}[h] %
  \centering
  \begin{subfigure}[h]{\linewidth}
    \centering
    \includegraphics[width=\linewidth]{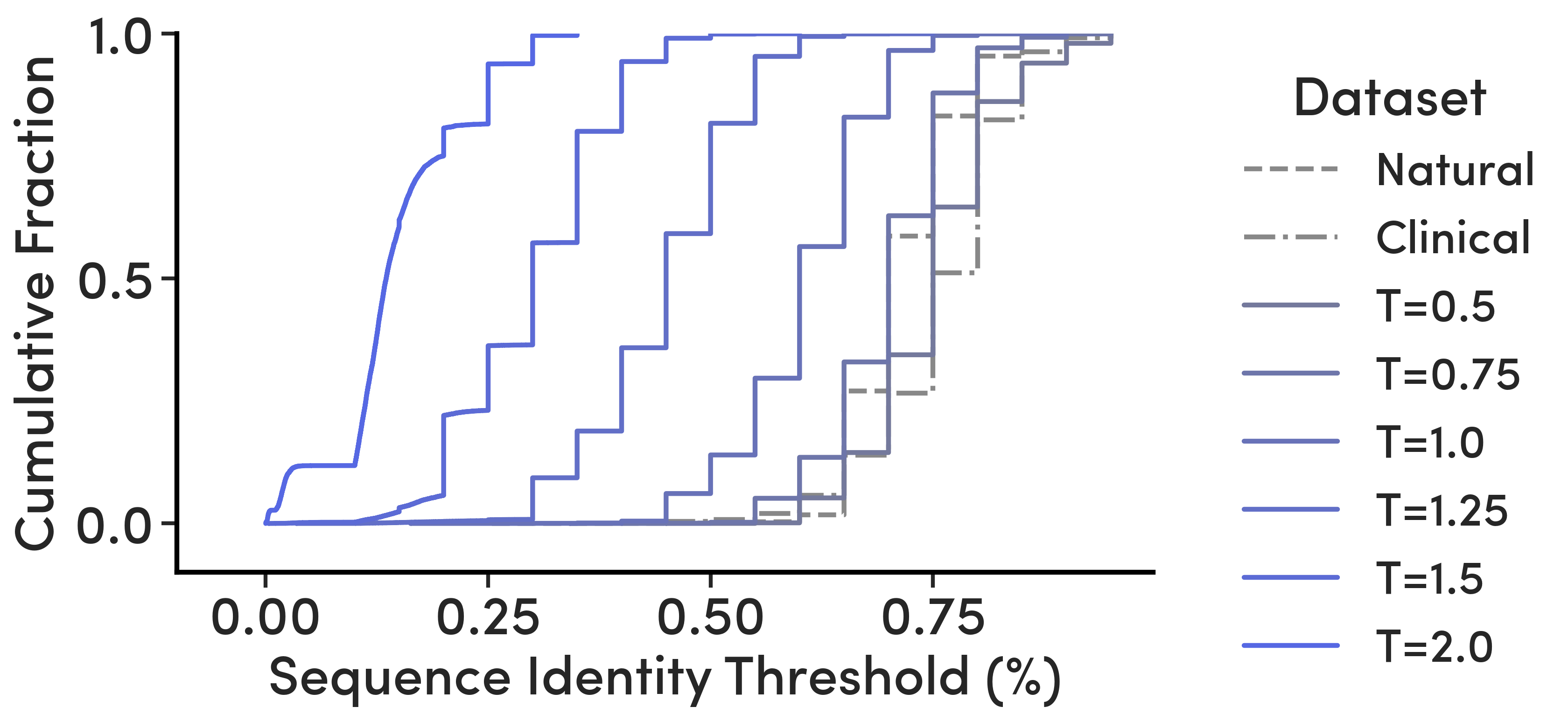}
    \caption{Empirical Cumulative Distribution Functions (ECDFs) of pairwise sequence similarities for Natural and Clinical antibodies, and sequences generated at different sampling temperatures.}
    \label{fig:generation_diversity}
  \end{subfigure}
  \begin{subfigure}[h]{\linewidth}
      \centering
      \includegraphics[width=\linewidth]{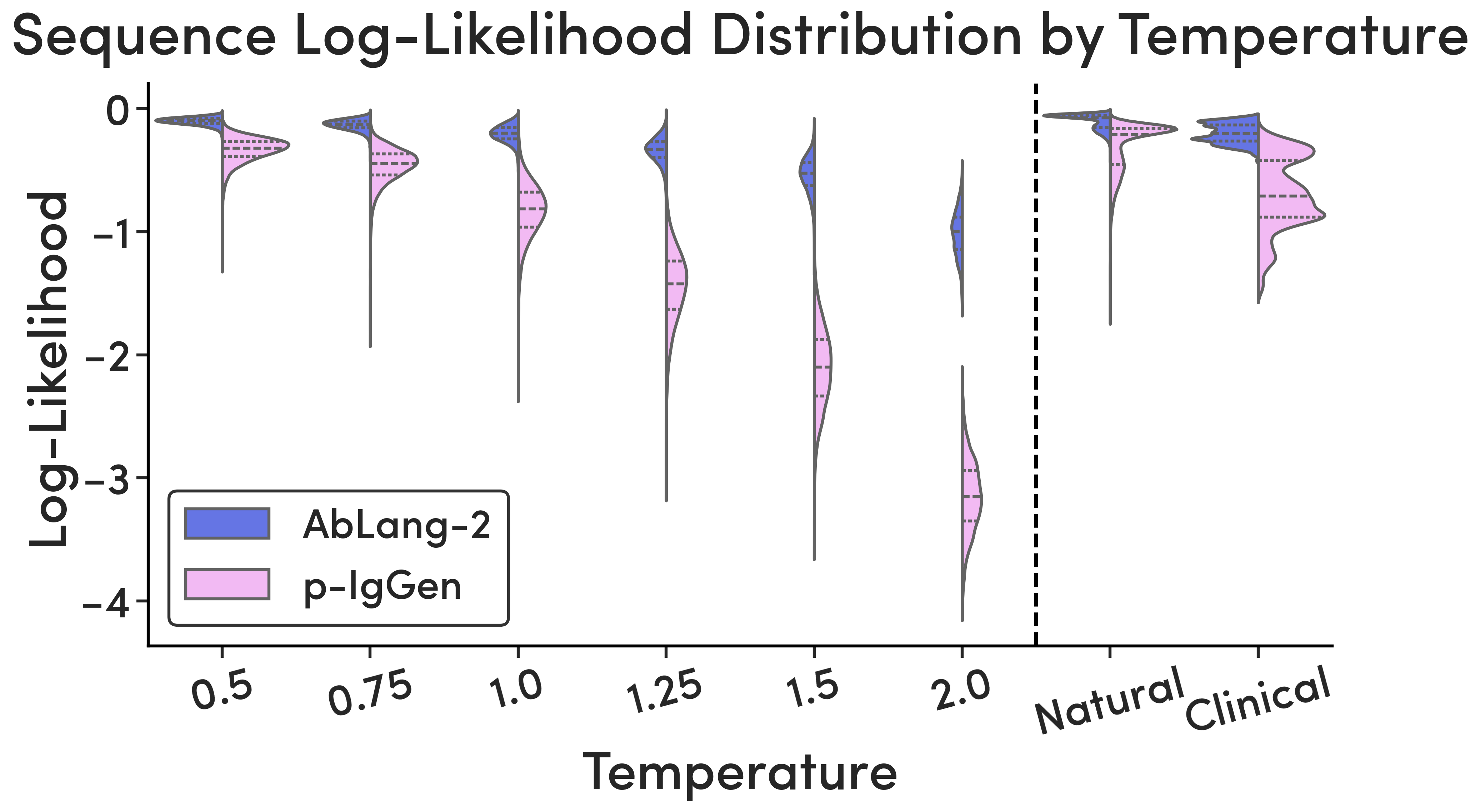}
      \caption{Violin plot of sequence naturalness scored with AbLang-2 and p-IgGen}
      \label{fig:generation_likelihoods}
  \end{subfigure}
    \caption{\textbf{Tunable generation of diverse antibodies.}}
    \label{fig:generation_figs}
\end{figure}

\subsection{Guided generation using the predictive models}

To bias sequence generation toward favorable developability characteristics, we  paired SVDD with our generative model. At each step, SVDD assesses several intermediate samples (“branches”) using dedicated scoring models and selects the one with the highest composite score. We approximate the SVDD value function via a posterior mean estimate (Figure \ref{fig:svdd_diagram}): for a given position, we sample multiple branches, fully denoise each, and compute scores by equally weighting the negative normalized AC-SINS and HIC measurements. The branch whose denoised sequence achieves the top score is retained as the current state, and the process repeats. This approach obviates the need to retrain predictors on partially masked data at varying masking levels and accommodates non-differentiable scoring functions.

%%%%%%%%%%%%%%%%%%%%%%%%%%%%%%%%%%%%%%%%%%%%%%%%%%%%
% Eval/results
%%%%%%%%%%%%%%%%%%%%%%%%%%%%%%%
\section{Results}

\begin{figure}[h]
  \centering
  \begin{subfigure}[b]{0.48\linewidth}
    \centering 
    \includegraphics[width=\linewidth]{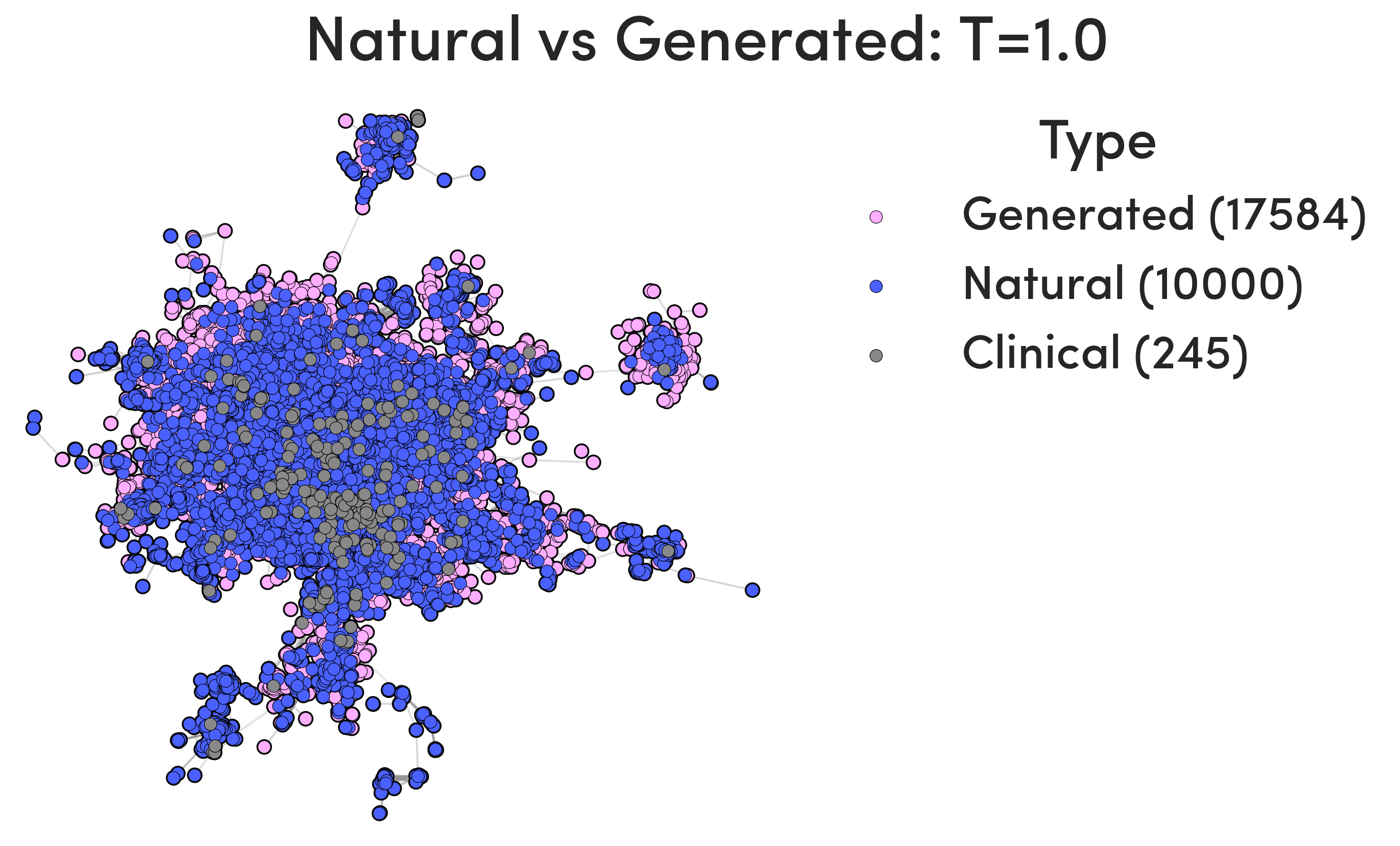}
    \caption{}
    \label{fig:unconditional_cluster_T1}
  \end{subfigure}%
  \hfill
  \begin{subfigure}[b]{0.48\linewidth}
    \centering
    \includegraphics[width=\linewidth]{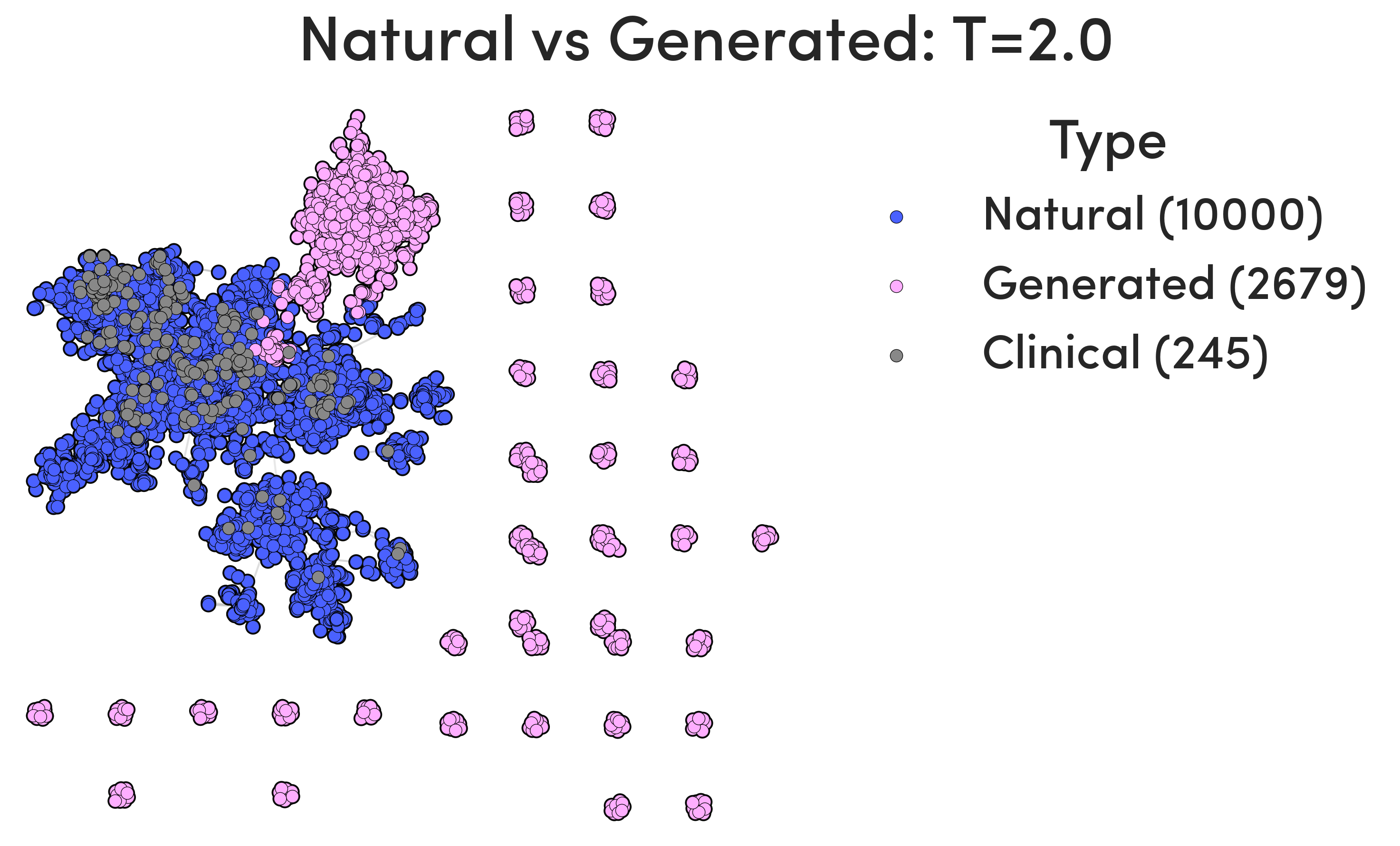}
    \caption{}
    \label{fig:unconditional_cluster_T2}
  \end{subfigure}
  \caption{\textbf{Sequence Similarity Networks of Generated Antibodies with Natural and Clinical Antibodies.} 
    a) The generated sequences cover a large region of natural sequence space, and clinical candidates are interspersed in that space.
    b) At high sampling temperatures, sequences are highly diverse but no longer overlap closely with natural and clinical sets. This also leads to many small disconnected clusters shown in pink.}
  \label{fig:unconditional_cluster}
\end{figure}
\subsection{Diversity/Naturalness}

We first evaluated our unconditional generation model’s ability to produce plausible antibodies (naturalness), and the diversity of the generated sequences. To assess naturalness, we scored generated sequences using two independent antibody-specific language models, AbLang2 \cite{Olsen2024-bh} and p-IgGen \cite{Turnbull2024-sn}, under different sampling temperatures. Both models are able to take paired heavy- and light-chain antibody sequences as input. To quantify diversity, we constructed sequence-similarity networks comparing generated sequences to 246 clinical antibodies in the training set, and to natural antibodies from the Observed Antibody Space (OAS) database, employing MMseqs2 \cite{Steinegger2017-rr}. We observed that higher sampling temperatures yield more diverse sequences (Figure~\ref{fig:generation_figs}a) at the expense of lower log-likelihood (naturalness) scores (Figure~\ref{fig:generation_figs}b). At $T = 1.0$, we found a favorable trade-off: 
generated sequences are diverse while retaining high log-likelihood scores (Figure~\ref{fig:generation_figs}a,b). This observation is also evident from the corresponding Sequence Similarity Network (SSN), and in contrast with sequences generated at T=2 (Figure~\ref{fig:unconditional_cluster}). % I like the cluster map to illustrate this point.
Interestingly, the natural and clinically approved antibodies are distributed throughout the sequence similarity network rather than being confined to tight clusters. This dispersion likely reflects the heterogeneous development trajectories of clinical candidates. Moreover, this sequence diversity in the training set of our predictive models, in addition to their robust cross-validation performance, supports that these models are well suited to guide antibody design.

\begin{figure}[ht]
    \centering
    \includegraphics[width=\linewidth]{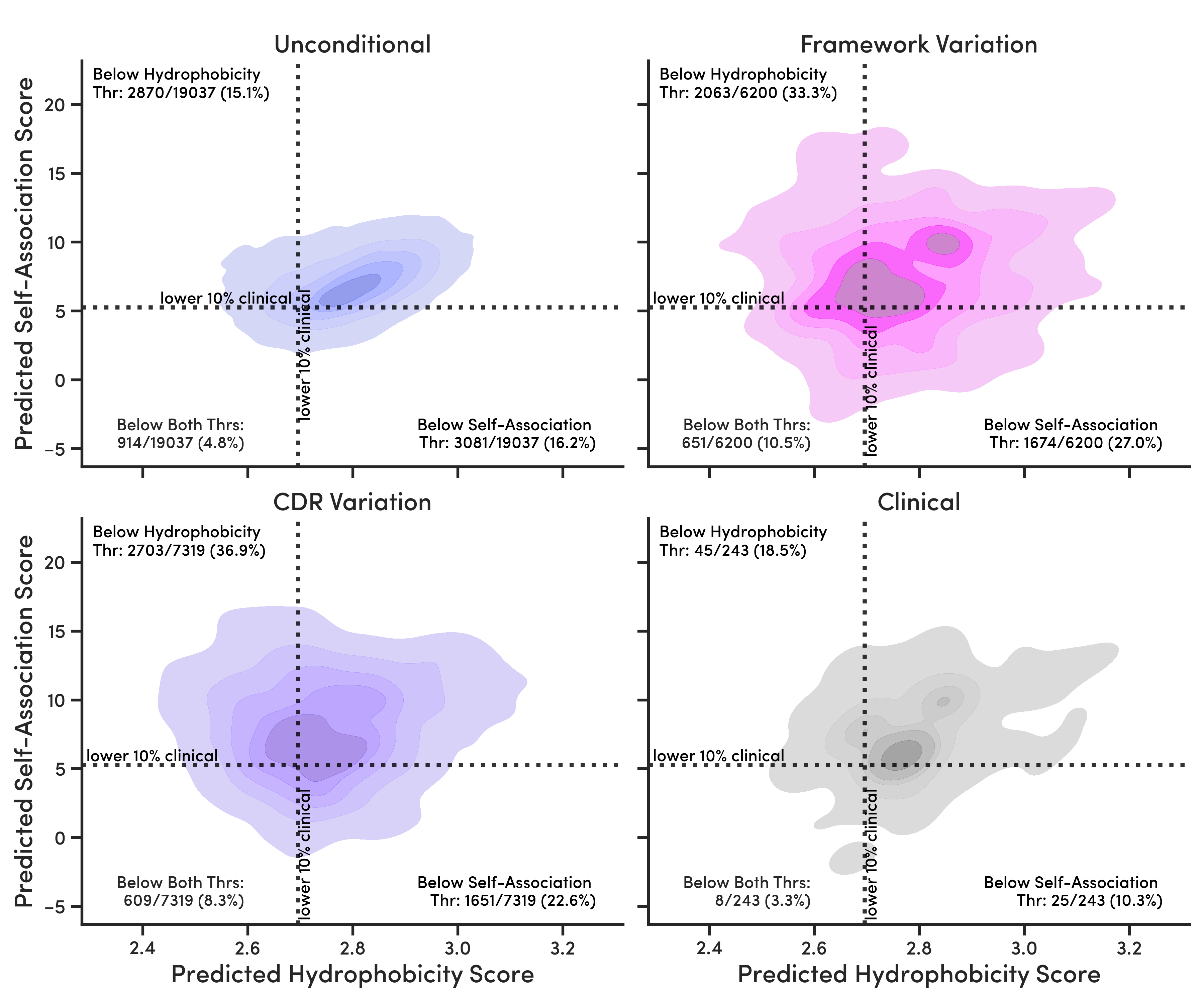}
    \caption{\textbf{Density plot of predicted self-association and hydrophobicity for various generation conditions.} Thresholds are calculated based on the lower 10\% of predicted properties of clinical antibodies. We show that we can increase the proportion of antibodies in the lower left quadrant (high developability) for both properties simultaneously using guided generation.}
    \label{fig:generated}
\end{figure}

\subsection{Guided Generation}
\begin{figure}[ht]
    \centering
    \includegraphics[width=\linewidth]{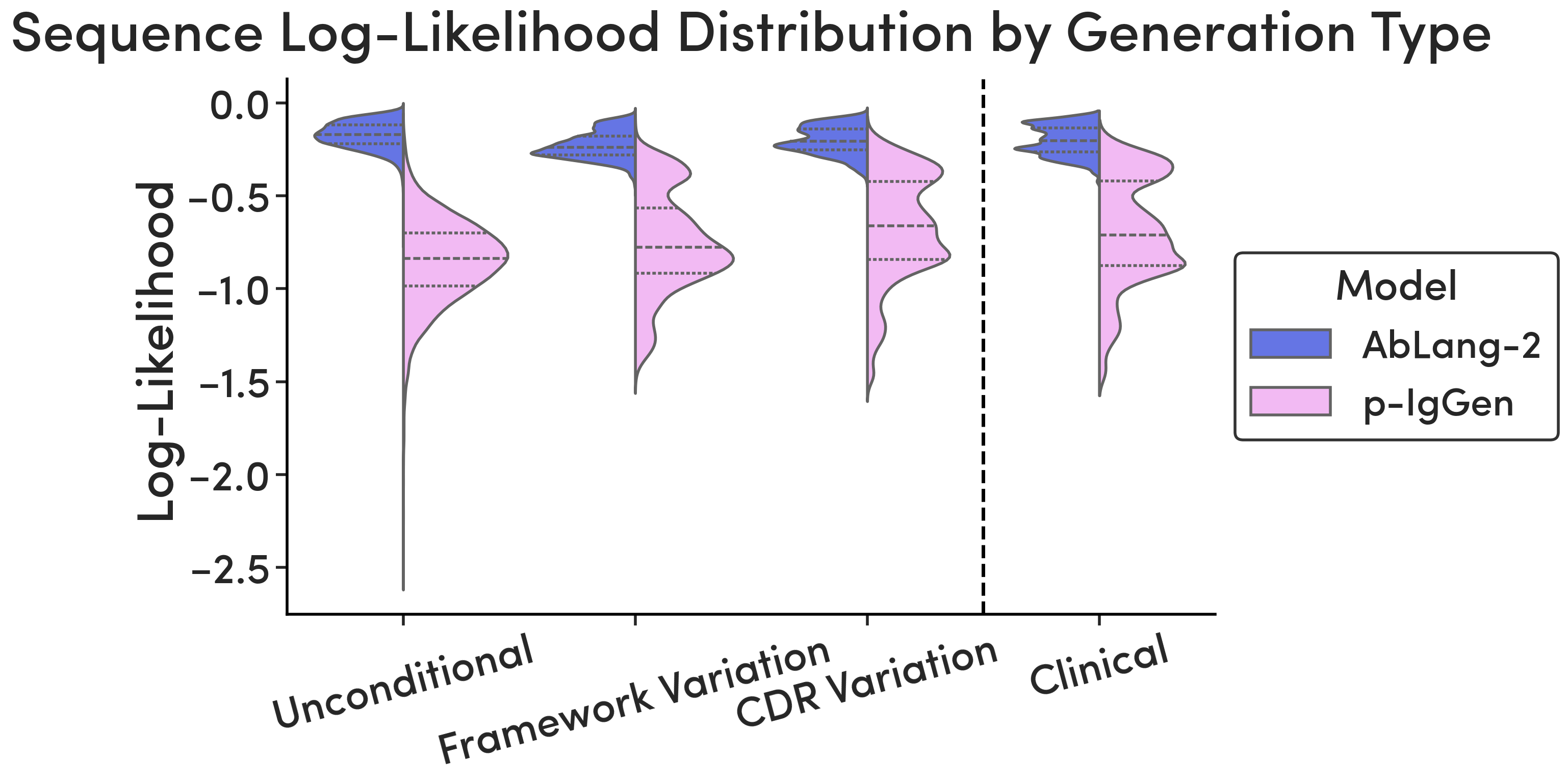}
    \caption{\textbf{Generated antibodies have high naturalness.} Violin plot of generated sequences scored with Ablang-2 and p-IgGen}
    \label{fig:generated_naturalness}
\end{figure}

We configured SVDD to guide generation using both hydrophobicity (HIC RT) and self-association (AC SINS) predictors. 
In total, we generated 13,519 antibody sequences with $T = 1.0$ across two experiments, masking the 1) Complementarity Determining Region of the Heavy chain (HCDR3), a highly variable sequence that influences developability and binding; and 2) the framework regions of both chains, more conserved sequences that provide a structural scaffold for CDRs.
We used the antibodies characterized in \cite{Arsiwala2025-fz} as starting template sequences for generation. 
Figure~\ref{fig:generated}a shows the joint distribution of hydrophobicity and self-association scores for the 1) unconditionally generated, 2) framework guided, 3) HCDR3 guided, and 4) the original clinical sequences used as templates. We indicate the 10th percentile of the measured clinical sequences as dotted line references. For both scores, our guided generation successfully enriched the number of sequences in the lower left quadrant, from 3.3\% in the seed sequence set, to 10.5\% and 8.3\% for framework and HCDR3 generation, respectively. These values were also higher than unconditional generation. In addition, generated guided sequences preserved the naturalness of unconditional and template clinical sequences (Figure \ref{fig:generated_naturalness}).

\section{Discussion and Future Works}

In this work, we demonstrate that discrete diffusion–based models can generate plausible and diverse antibody sequences. We find that while high sampling temperatures increase diversity, they also degrade naturalness. When coupled with simple biophysical predictors, these models can effectively explore novel regions of therapeutic sequence space while preserving naturalness.
A natural next step is developing better predictors for biophysical properties. 
Many antibody biophysical property predictors have been developed~\cite{Jain2017-wj,Tomar2017-vn, Khurana2018-ek,Thumuluri2021-iy,Zhou2022-ta,Lai2022-pi,Gentiluomo2020-gm,Prihoda2022-gy,Schmitt2023-sp,Wu2025-rx,Park2024-dy,Rollins2024-ze} but restrictive commercial terms, difficulty installing, inconsistent benchmark comparisons or insufficient performance limit their use. 
Additionally, developing predictors across highly diverse sequences has been found to be much more challenging than predicting local mutational effects~\cite{Groth2023-sr,Notin2023-ta}, but with larger amounts of high quality training data we are confident that generalized predictors across natural and clinical antibody space (and eventually, across different formats such as Fc-fusions, single domain antibodies (VHH, scFv, monobodies), and multispecifics) could achieve much higher accuracy and hence improve guided generation quality.

To advance this framework, we plan to incorporate larger, more diverse labeled datasets covering additional biophysical assays and to investigate multi-task predictors that jointly model target binding as well as developability. Extending our guidance beyond posterior-mean SVDD may enable stronger steering with fewer denoising steps. We will also validate generated candidates experimentally to quantify real-world developability gains and refine our predictors accordingly.

% \section*{Acknowledgements}

% \textbf{Do not} include acknowledgements in the initial version of
% the paper submitted for blind review.

% If a paper is accepted, the final camera-ready version can (and
% usually should) include acknowledgements.  Such acknowledgements
% should be placed at the end of the section, in an unnumbered section
% that does not count towards the paper page limit. Typically, this will 
% include thanks to reviewers who gave useful comments, to colleagues 
% who contributed to the ideas, and to funding agencies and corporate 
% sponsors that provided financial support.

\section*{Impact Statement}
Antibodies comprise a major modality of therapeutics, and have been used to treat cancer, autoimmune diseases and infectious diseases. We expect that machine learning research focused on optimizing developability will have a positive social impact in reducing clinical trial costs and enabling more therapies to come to market.

% Antibodies comprise a major modality of therapeutics. Traditional framework of antibody development 
% Authors are \textbf{required} to include a statement of the potential 
% broader impact of their work, including its ethical aspects and future 
% societal consequences. This statement should be in an unnumbered 
% section at the end of the paper (co-located with Acknowledgements -- 
% the two may appear in either order, but both must be before References), 
% and does not count toward the paper page limit. In many cases, where 
% the ethical impacts and expected societal implications are those that 
% are well established when advancing the field of Machine Learning, 
% substantial discussion is not required, and a simple statement such 
% as the following will suffice:

% ``This paper presents work whose goal is to advance the field of 
% Machine Learning. There are many potential societal consequences 
% of our work, none which we feel must be specifically highlighted here.''

% The above statement can be used verbatim in such cases, but we 
% encourage authors to think about whether there is content which does 
% warrant further discussion, as this statement will be apparent if the 
% paper is later flagged for ethics review.

\clearpage
\bibliography{paperpile}
\bibliographystyle{icml2025}

%%%%%%%%%%%%%%%%%%%%%%%%%%%%%%%%%%%%%%%%%%%%%%%%%%%%%%%%%%%%%%%%%%%%%%%%%%%%%%%
%%%%%%%%%%%%%%%%%%%%%%%%%%%%%%%%%%%%%%%%%%%%%%%%%%%%%%%%%%%%%%%%%%%%%%%%%%%%%%%
% APPENDIX
%%%%%%%%%%%%%%%%%%%%%%%%%%%%%%%%%%%%%%%%%%%%%%%%%%%%%%%%%%%%%%%%%%%%%%%%%%%%%%%
%%%%%%%%%%%%%%%%%%%%%%%%%%%%%%%%%%%%%%%%%%%%%%%%%%%%%%%%%%%%%%%%%%%%%%%%%%%%%%%
% \newpage
% \appendix
% \onecolumn

% You can have as much text here as you want. The main body must be at most $8$ pages long.
% For the final version, one more page can be added.
% If you want, you can use an appendix like this one.  

% The $\mathtt{\backslash onecolumn}$ command above can be kept in place if you prefer a one-column appendix, or can be removed if you prefer a two-column appendix.  Apart from this possible change, the style (font size, spacing, margins, page numbering, etc.) should be kept the same as the main body.
%%%%%%%%%%%%%%%%%%%%%%%%%%%%%%%%%%%%%%%%%%%%%%%%%%%%%%%%%%%%%%%%%%%%%%%%%%%%%%%
%%%%%%%%%%%%%%%%%%%%%%%%%%%%%%%%%%%%%%%%%%%%%%%%%%%%%%%%%%%%%%%%%%%%%%%%%%%%%%%

\end{document}